# Use GPT-J Prompt Generation with RoBERTa for NER Models on Diagnosis Extraction of Periodontal Diagnosis from Electronic Dental Records

Yao-Shun Chuang, MS[1], Xiaoqian Jiang, PhD[1], Chun-Teh Lee, DDS, MS, DMSc[2], Ryan Brandon, MS[3], Duong Tran, PhD[5], Oluwabunmi Tokede, DDS, MPH, DMSc[4,5], Muhammad F. Walji, PhD[4,5]

[1]McWilliams School of Biomedical Informatics, University of Texas Health Science Center at Houston, Houston, Texas, USA; [2]Department of Periodontics and Dental Hygiene, The University of Texas Health Science Center at Houston School of Dentistry, Houston, Texas, USA; [3]Department of Oral Health Sciences, Temple University Kornberg School of Dentistry, Philadelphia, Pennsylvania, USA; [4]Oral Healthcare Quality and Safety, The University of Texas Health Science Center at Houston School of Dentistry, Houston, Texas, USA; [5]Diagnostic and Biomedical Sciences, The University of Texas Health Science Center at Houston School of Dentistry, Houston, Texas, USA

**Abstract**
*This study explored the usability of prompt generation on named entity recognition (NER) tasks and the performance in different settings of the prompt. The prompt generation by GPT-J models was utilized to directly test the gold standard as well as to generate the seed and further fed to the RoBERTa model with the spaCy package. In the direct test, a lower ratio of negative examples with higher numbers of examples in prompt achieved the best results with a F1 score of 0.72. The performance revealed consistency, 0.92-0.97 in the F1 score, in all settings after training with the RoBERTa model. The study highlighted the importance of seed quality rather than quantity in feeding NER models. This research reports on an efficient and accurate way to mine clinical notes for periodontal diagnoses, allowing researchers to easily and quickly build a NER model with the prompt generation approach.*

**Introduction**

Periodontitis is a common dental disease that affects almost half of adults aged 30 or older in the United States [1]. It is characterized by gingival inflammation and bone loss around teeth, and if left untreated, can significantly reduce the quality of life [1,2]. According to a US national surveillance project, 42% of dentate adults aged 30 and above were affected by periodontitis, with 7.8% experiencing severe conditions [1]. In 2018, a new classification system for periodontal diseases was introduced, which includes staging, extent, and grading to diagnose periodontitis [3]. The determination of the disease stage is based on the severity of the disease during its presentation and the complexity of its management. The extent is indicated by the percentage of teeth affected by periodontitis at the identified stage. Grading depends on the risk of disease progression associated with the history of disease progression, local and systemic factors. Despite the introduction of new diagnostic terms for periodontal diseases, dental care providers might not be acquainted with them due to the complexity of this new system. This results in clinical documentation lacking accurate and structured diagnosis, or in some cases, no diagnosis being recorded. Inadequate periodontal diagnoses poses a significant threat to patient care quality. An accurate diagnosis is key to the provision of appropriate patient care, outcome assessment and quality improvement efforts. This, in turn, may hinder future care providers from evaluating the patient's condition precisely and providing optimal treatment.

Electronic dental records (EDR) have become widely adopted in dental care, providing an opportunity to address the issue of missing diagnoses. EDRs include comprehensive information on a patient's history, clinical examination, diagnosis, treatment, and prognosis [4]. Maintaining accurate and detailed records is crucial for delivering high-quality patient care and ensuring appropriate follow-up [4]. Thus, EDRs can be a reliable source of clinical information, but much of this information is often documented in an unstructured free-text format. Dental care providers often write detailed treatment plans and prognostic factors in free-text format for clinical care purposes. [5] While this information is easily accessible during patient care, generating meaningful insights for secondary analysis can be challenging.[6] The unstructured nature of these records requires manual review by domain experts, which can be time-consuming and costly, particularly when dealing with a large number of patient records.[7] Proper management and analysis of dental records are necessary to derive meaningful information.[8]

To overcome this challenge, natural language processing (NLP) techniques can be applied to extract structured data from unstructured EDRs. Named Entity Recognition (NER) is a key NLP task that involves identifying and extracting specific entities of interest, such as disease diagnoses, medication names, and lab tests, from clinical narratives [9]. The NLP approaches in the last decade mainly focused on supervised, semi-supervised, and unsupervised NER systems [10]. Due to the rapid advancement of hardware and technology, neural network NER systems have become more prevalent in recent years [11]. After surveying 23 papers published after 2016, a review article on clinical NER reported that machine learning-based methods were conducted in 15 articles, rule-based methods in 5 papers, and language model base methods in 4 papers. Notably, only 2 papers out of 23 papers used hybrid approaches, which combine more than one method [12]. All of these studies demonstrated exceptional performance in NER tasks. However, the majority of these models lacked generalizability and take time to build due to their dependence on data.[12] Therefore, by leveraging NLP techniques and large language models (LLM)[13] like the GPT series, this study analyzes the usability of the prompt generation from the large language model to create the seed and further train by BERT[14] models.

To avoid privacy concerns, we employed GPT-J[15] (which can be deployed locally) to generate prompts, while RoBERTa[16] is utilized for NER modeling. Moreover, the study examined various prompt settings to determine their effectiveness. The study presented a pipeline that utilizes prompt generation to create a quality seed and train on a NER model. This approach offers a simpler solution for seed creation and demonstrates a high potential for expansion.

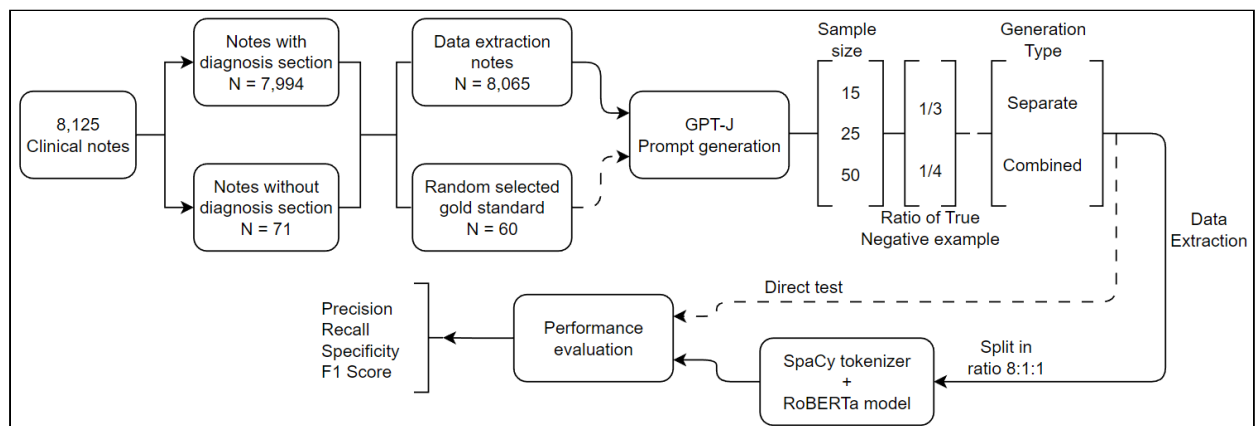

**Figure 1.** A flowchart of this study from data collection, seed generation through prompt and then NER model training for named entity recognition.

**Methods**

*Dataset*

The data for this study were obtained from an EDR covering the time frame of January 1, 2021 to December 31, 2021. To be included in the study, cases had to have undergone a comprehensive periodontal examination that included pocket depths, clinical attachment loss, and measurements of the free gingival margin to the cemento-enamel junction. Additionally, cases were restricted to individuals over the age of 16 with at least 10 natural teeth and had undergone bitewing radiographs within 6 months of the examination – all these information are critical for arriving at an accurate periodontal diagnosis. The dataset contained a total of 5,495 eligible patients who met the specified criteria. Clinical case notes from within one month of the examination were extracted for each of these cases, resulting in a total of 8,125 clinical notes. To verify the accuracy of the model's predictions, 60 clinical notes were randomly selected and manually labeled by an experienced dentist who had not written any of the notes. These labeled notes were excluded from the training dataset for the NLP model. Therefore, the final training dataset consisted of 8,065 clinical notes.

*Target information*

This study used the clinical notes of patients diagnosed with any periodontal diagnosis , in accordance with the American Academy of Periodontology (AAP)/European Federation of Periodontology (EFP) 2018 classification system [17]. In the diagnosis of periodontitis, entities include four stages (I, II, III, and IV), three grades (A, B, C), and three types of extent (localized, generalized, and molar/incisor pattern). However, the molar/incisor pattern extent was not taken into account in this study due to its infrequency. In addition, only periodontitis diagnoses were assessed in the study.

*NLP building blocks*

We leveraged the capabilities of the spaCy package for tokenization and the integration of RoBERTa base into NER models, as well as the powerful GPT-J-6B transformer model for seed generation. We explored the performance of these models in the extraction of diagnoses from clinical notes. Our approach included a rule-based method for diagnosis extraction, followed by seed generation using GPT-J-6B. The results were then fine-tuned to improve NER model performance.

Tokenization: The spaCy package, a widely-used and efficient library for natural language processing, was utilized for tokenization. Its non-destructive approach ensures the preservation of whitespace and punctuation, which is crucial for maintaining the original structure of clinical notes. The package's architecture allows users to customize the NLP pipeline, enabling the addition or removal of specific components to suit the task at hand. This flexibility contributes to high performance and adaptability in various NLP tasks. Once the data was tokenized, it was reconstructed and saved in the spaCy training data format, facilitating seamless integration with other NLP tools and models.

RoBERTa Integration for Enhanced NER Performance: RoBERTa base, a powerful and efficient pre-trained transformer model, was effectively integrated into the NER models in this study. Building on the benefits offered by spaCy tokenization, RoBERTa base significantly bolsters the NER model's ability to detect and categorize named entities within clinical notes. The integration process entailed fine-tuning RoBERTa base with the restructured spaCy training data and harnessing its contextual embeddings to augment the performance of the NER models. RoBERTa is an advanced version of the BERT architecture that employs a dynamic masked language modeling technique during pretraining to circumvent over-memorization of the training data.[18] This feature renders it a superior option for general-purpose tasks, such as the extraction of target information in this study.[16]

GPT-J-6B Transformer Model: The GPT-J-6B model is one of the most powerful and easily usable transformer models and Large Language Models (LLMs) in the world. The model has 28 layers and a model dimension of 4096, with 16 heads each having a dimension of 256. Rotary position encodings (RoPE) were used on 64 dimensions of each head. The model was trained with a tokenization vocabulary of 50257 using the same set of BPEs as GPT-2/GPT-3[19]. Like other GPT series, these state-of-art LLMs are highly effective at a variety of tasks, including text generation, sentiment analysis, classification, and machine translation[20]. During the development process, GPT-J-6B was selected and employed on the server because of hardware considerations and data privacy issues. Most LLMs are either too large to download or are only available as cloud-based services. For example, ChatGPT, the most popular LLM, fits both situations. However, the clinical data contains protected health information (PHI) that is regulated by Health Insurance Portability and Accountability Act (HIPAA). Despite ChatGPT's robust language generation capabilities, its use of cloud-based services can lead to the collection and temporary storage of users' information in the system. This could potentially result in privacy data leakage and a violation of HIPAA regulations. Thus, the GPT-J-6B model was obtained and utilized on a local server.

*Seed Generation using LLM for NER*

The seed generation utilized GPT-J-6B to prompt the target information and **Figure 1** demonstrated the workflow of this study. The rule-based method was first applied to the clinical notes to extract diagnosis sentences and randomly selected samples for the prompt examples. These examples were mixed with positive and negative examples.

Rule-based Diagnosis Extraction: A rule-based method was first applied to the clinical notes to extract diagnosis sentences. This method involved searching for specific keywords, phrases, or patterns indicative of diagnoses within the text. Once identified, randomly selected samples were chosen for prompt examples to train the GPT-J-6B model in generating seeds for further analysis.

Mixing Positive and Negative Examples: To create a balanced dataset for training, positive and negative examples were combined. Positive examples contained details of a periodontitis diagnosis, while negative examples did not. In this study, the first sentence in **Table 1** served as a negative example, and sentences No. 2-4 functioned as positive examples.

**Table 1.** Example of prompt generation format.

| No. | Example sentences | Prompt | Results |
|---|---|---|---|
| 1 | d: patient presents for periodic oral examination and prophy | Stage/Grade/Extent: | None/None/None |
| 2 | d: localized stage ii grade b periodontitis | Stage/Grade/Extent: | ii/b/localized |
| 3 | d: tentaive diagnosis: perio stage iii grade b, confirm needed. | Stage/Grade/Extent: | iii/b/None |
| 4 | d: generalized stage iii periodontitis, grade b. | Stage/Grade/Extent: | iii/b/generalized |
| 5 | d: generalized chronic periodontitis stage 3 grade b | Stage/Grade/Extent: | |
| 6 | d: tentaive diagnosis: perio stage iii grade b, confirm needed. | Stage: | iii |
| 7 | d: generalized chronic periodontitis stage 3 grade b | Stage: | |

Prompt Engineering: To optimize the performance of the GPT-J-6B model during seed generation, a crafted prompt section was manually appended to the end of each input sentence. This prompt section included three distinct labels separated by a slash mark, along with their corresponding results. This specific format not only provided the model with clear and concise guidance but also allowed it to effectively learn the desired output structure for each named entity. By incorporating these tailored prompts, the GPT-J-6B model could better understand the context and produce more accurate and relevant results, ultimately enhancing the overall effectiveness of the AI-driven solution. In this manner, prompt engineering plays a crucial role in streamlining the learning process and improving the performance of the language model in various applications. The examples of a combined and separate type of generation were illustrated in **Figures 2** and **3**, respectively.

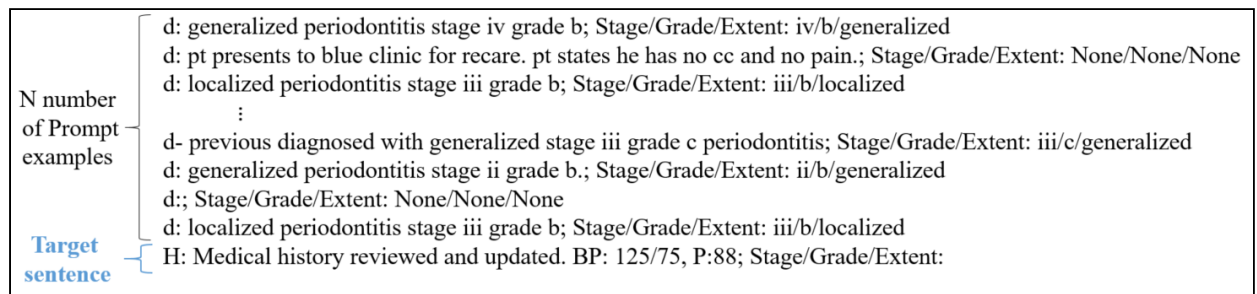

**Figure 2.** A prompt example of the combined type generation. N was the number of example sentences used.

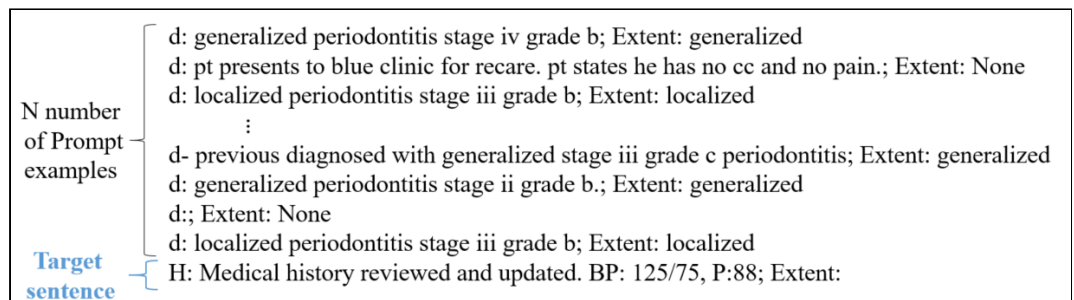

**Figure 3.** A prompt example of the separate type generation. N was the number of example sentences used.

Sample Size, Negative Example Ratios, and Generation Types: To evaluate the impact of various factors on model performance, the study tested three sample sizes (15, 25, and 50), two ratios of negative examples (one-third and one-quarter of the sample size), and two generation types (separate and combined). The fifth sentence in **Table 1** was represented as the target sentence. The entire clinical notes were split by the newline character and each sentence was appended at the end of example sentences followed by the prompt for GPT-J to produce the results. The examples in No. 1-5 in **Table 1** were the combined generation, whereas the separate generation (No. 6-7) prompted each label individually to get the result and then combined all three labels together for that target sentence. This experimentation helped identify optimal configurations for generating high-quality seeds using the GPT-J-6B model.

*Seed Splitting for Fine-tuning*

The individual sentences in the clinical notes were separated using the newline character and the prompt section was then applied to generate seeds for named entities. The results from the GPT-J-6B generated prompt approach were split in a ratio of 8:1:1 for fine-tuning the NER model. This partition allocated 80% of the data for training, 10% for validation, and the remaining 10% for testing. **Table 2** provides a detailed breakdown of the data distribution for each subset. By fine-tuning the NER model on the generated seed data, the model's performance was optimized to better recognize and classify named entities in clinical notes. The iterative process of training, validation, and testing ensured that the model generalized well and avoided overfitting on the training data. This fine-tuning step was crucial in enhancing the NER model's overall accuracy and effectiveness in extracting diagnostic information from clinical notes.

**Table 2.** Distribution of extraction results from the prompt approach in the types of combined generation.

|  | Sample size | Total Sentences Extraction | Train | Validation | Test |
|---|---|---|---|---|---|
| Ratio (1/3) | 15 | 151 | 120 | 15 | 16 |
|  | 25 | 108 | 86 | 11 | 11 |
|  | 50 | 102 | 81 | 10 | 11 |
| Ratio (1/4) | 15 | 517 | 413 | 52 | 52 |
|  | 25 | 270 | 216 | 27 | 27 |
|  | 50 | 31 | 24 | 3 | 4 |

*Data Pre-processing and Post-processing Enhancements*

We implemented additional data pre-processing and post-processing steps to ensure the quality of the results. In the pre-processing phase, to address the issue of AI hallucination, labels were extracted from the prompt and verified for their presence in the target sentence for both types of generation. This procedure guaranteed that the target labels and data were valid for NER model training. Following the generation of results from the NER model, data underwent post-processing to ensure proper generalization. While the information was accurately extracted from the notes, it might not have been suitable for evaluation purposes without further refinement. Post-processing steps included rectifying typographical errors in the Extent and standardizing Stage and Grade values. Some Grade labels contained extraneous symbols, such as dots and commas, which were eliminated during post-processing. If the data could not be generalized, the value was left blank. Since a single clinical note might contain multiple diagnoses, the most severe diagnosis was chosen by comparing the severity order in the following hierarchy: Stage, Extent, and then Grade.

*Evaluation metrics*

To evaluate the performance of the NER model, a confusion matrix was employed. The true positive (TP) values indicate instances are recognized as the presence of periodontitis by the NER model correctly based on the gold standard. Conversely, true negative (TN) values indicate instances where neither the gold standard nor the NER model identifies the presence of the disease. False positive (FP) values represent cases where the NER model incorrectly predicts the presence of the disease, and false negative (FN) values indicate instances where the NER model fails to identify the actual presence of periodontitis. To assess the algorithm's performance, six metrics were used. Precision (P), also known as a positive predictive value, measures the proportion of true positives out of all positive predictions. Recall, also known as true positive rate or sensitivity, measures the proportion of true positives out of all actual positive cases. Specificity tests the ratio of true negatives to all negative predictions. The F1 score is the harmonic mean of precision and recall. Due to the uncertainty and imbalance of the distribution of disease status in periodontitis, the macro average and weighted average were used for the metrics mentioned above. The macro average is calculated by averaging the evaluation values in the respective fields, while the weighted average is generated by weighing the evaluation values by their corresponding quantity.

**Results**

First, we assessed the few-shot performance of the results generated by the seeds. The GPT-J-6B model's immediate outputs were evaluated against 60 gold standard clinical notes and subsequently used to train the NER model for comparison with the gold standard. The results of this direct test are illustrated in **Figure 4**. **Figure 4(a)** displays the performance when training the model on one-third negative examples, with precision for Stage, Grade, and Extent ranging from 0.45 to 0.82 in the macro average and 0.51 to 0.85 in the weighted average across sample sizes of 15, 25, and 50. Both recall and F1 scores were below 0.46. In contrast, **Figure 4(b)** presents the model trained with samples of one-quarter negatives, demonstrating that precision for the three labels across the three sample sizes varied from 0.56 to 0.88 in both macro and weighted averages. The recall and F1 scores ranged from 0.45 to 0.72. Compared to the results from the one-third negative ratio, the variance in precision was below 0.10. However, the recall demonstrated an increase of 0.11, 0.23, and 0.23, while the F1 score exhibited an increase of 0.11, 0.28, and 0.27 in sample sizes of 15, 25, and 50, respectively.

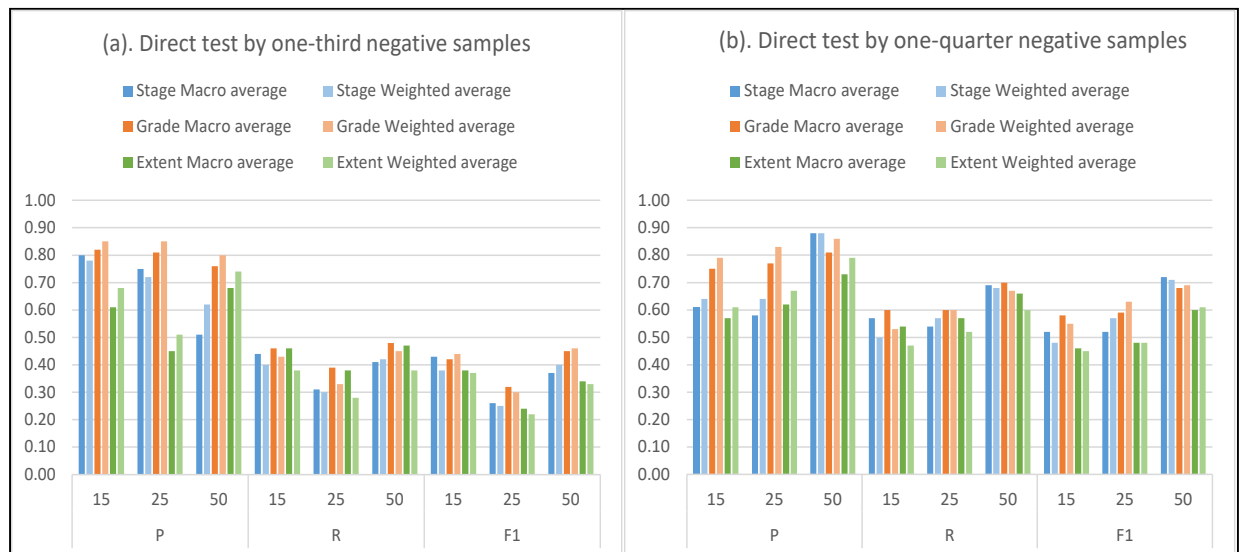

**Figure 4.** The bar charts of prompt results by direct testing on the gold standard.

Subsequently, we utilized the seeds generated by both prompted results as input for the RoBERTa model to conduct the final classification evaluation. By feeding these seeds to the RoBERTa model, we aimed to investigate the effectiveness of the generated seeds in training the NER model and to compare the performance metrics across different prompt engineering scenarios. **Table 3** and **4** show the evaluation metrics of the NER model predictions in three sample sizes from the ratio of one-third and one-quarter types, respectively.

**Table 3** compares the performance of RoBERTa models trained by one-third negatives with different sample sizes (15, 25, and 50) across precision (P), recall (R), and F1 score (F1). The table presented results for macro and weighted averages in three categories: Stage, Grade, and Extent. For the Stage and Grade categories, both macro and weighted averages demonstrated high performance, with scores generally ranging from 0.95 to 0.98. The Extent category shows slightly lower scores, with macro and weighted averages fluctuating between 0.89 and 0.92. This indicates that RoBERTa models exhibit strong performance across varying sample sizes, with some variations in the Extent category.

**Table 3.** The comparison between RoBERTa models trained by one-third negatives with different sample sizes.

|  |  | P | | | R | | | F1 | | |
|---|---|---|---|---|---|---|---|---|---|---|
|  | *Sample size* | *15* | *25* | *50* | *15* | *25* | *50* | *15* | *25* | *50* |
| **Stage** | Macro average | 0.95 | 0.98 | 0.97 | 0.95 | 0.97 | 0.96 | 0.95 | 0.97 | 0.96 |
|  | Weighted average | 0.97 | 0.98 | 0.97 | 0.97 | 0.98 | 0.97 | 0.97 | 0.98 | 0.97 |
| **Grade** | Macro average | 0.96 | 0.98 | 0.96 | 0.96 | 0.98 | 0.97 | 0.96 | 0.98 | 0.97 |
|  | Weighted average | 0.97 | 0.98 | 0.97 | 0.97 | 0.98 | 0.97 | 0.97 | 0.98 | 0.97 |
| **Extent** | Macro average | 0.92 | 0.92 | 0.89 | 0.90 | 0.90 | 0.89 | 0.91 | 0.91 | 0.89 |
|  | Weighted average | 0.92 | 0.92 | 0.90 | 0.92 | 0.92 | 0.90 | 0.92 | 0.92 | 0.90 |

Similarly, **Table 4** presented a comparison of RoBERTa models trained using one-quarter negatives with different sample sizes like in the previous table. Across all sample sizes, the macro and weighted averages for Stage and Grade remain consistently high, with scores around 0.95-0.97. In the Extent category, the macro and weighted averages show a slight variation, ranging from 0.86 to 0.92. Comparing the results of this model trained with one-quarter negatives, the model training by one-third negative was 0.01-0.02 better but the difference is small. This shows that the RoBERTa models exhibit strong performance and robustness in terms of seeds, indicating their effectiveness in this application.

**Table 4.** The comparison between RoBERTa models trained by one-quarter negatives with different sample sizes.

|  |  | P | | | R | | | F1 | | |
| --- | --- | --- | --- | --- | --- | --- | --- | --- | --- | --- |
| | Sample size | *15* | *25* | *50* | *15* | *25* | *50* | *15* | *25* | *50* |
| **Stage** | Macro average | 0.97 | 0.95 | 0.95 | 0.95 | 0.95 | 0.95 | 0.96 | 0.95 | 0.95 |
| | Weighted average | 0.97 | 0.97 | 0.97 | 0.97 | 0.97 | 0.97 | 0.97 | 0.97 | 0.97 |
| **Grade** | Macro average | 0.96 | 0.96 | 0.96 | 0.96 | 0.96 | 0.96 | 0.96 | 0.96 | 0.96 |
| | Weighted average | 0.97 | 0.97 | 0.97 | 0.97 | 0.97 | 0.97 | 0.97 | 0.97 | 0.97 |
| **Extent** | Macro average | 0.88 | 0.90 | 0.92 | 0.88 | 0.86 | 0.90 | 0.88 | 0.87 | 0.91 |
| | Weighted average | 0.88 | 0.89 | 0.92 | 0.88 | 0.88 | 0.92 | 0.88 | 0.88 | 0.92 |

**Discussion**

This study found that prompt generation could not only generate a meaningful seed but also provide a quick and easy approach to training the NER models. The direct use of the prompt generation on the NER prediction was also another possible approach although there was no useful data generated from the separate type of generation. There weren't any matched sentences captured in any of the prompt generations in any individual labels. Furthermore, the AI hallucination occurred in a separate type of generation, in which sometimes the generated prompts didn't exist in the original sentences. In contrast, the combined generation would be able to produce feasible and useful results. For the settings of the prompt on the direct test scenario, a better performance was revealed with a lower ratio of negative examples. Additionally, the precision, recall, and F1-score showed higher accuracy with more examples used in the prompt though the number of extracted sentences was decreased.

The findings of this study indicated that prompt generation narrowed the gap of seed generation, compared with regular expression or rule-based methods, and guaranteed promising training data for the NER models. The conventional approaches by regular expression, rule-based, dictionary-pair based methods requires the developer not only obtain domain knowledge but also manually go through tens to hundreds of data to create comprehensive algorithms to cover all the use cases. The manual process is time consuming and costly and is less likely to cover all the scenarios. Also, the built models are commonly less generalizable and still not perfect due to data dependency. Thus, prompt generation proved to be an effective method for producing training seeds for NER models. Furthermore, this pipeline's rapid adaptability provides significant extensibility, enabling the model to incorporate other dental conditions, in addition to periodontitis, quickly. This integration results in the generation of comprehensive and detailed structured data in electronic dental records, which enhance clinical applications for research and quality improvement in particular.

The learning capability of NER models was exemplified in various prompt generation scenarios, highlighting their strengths. However, direct use cases resulted in high false negatives, which were resolved through training with the RoBERTa model. Notably, the training samples ranged from 517 sentences to 31 sentences, indicating significant diversity. Following the training process of NER models, the accuracy was consistent across all settings. This outcome suggested that the quality of the seed used to feed the NER models was critical, rather than the quantity. Moreover, the recall metrics were mostly lower than precision, which is desirable since the objective is to identify missed diagnoses. Upon conducting an error analysis, it was found that in around 10 cases out of the 60 gold standard notes, the models were able to correctly extract the target information, but some marks like dots, commas, and other symbols were included. This led to errors in the evaluation metrics, indicating the need for post-processing utilizing rule-based text processing after obtaining the model's prediction. To ensure data standardization, post-processing was also essential since the NER model had correctly identified certain target terms that contained typos and informal language, such as "Grade D" being written as "Grade iii" and other incorrect values. Therefore, post-processing is necessary to rectify these issues.

One notable limitation of this study was the significant hardware requirements. The seed generation processes for all settings, involving executing 10 times on over 8,000 notes, took over a week to complete using 3 NVIDIA A100 GPUs. Despite these challenges, the benefits of this experiment are expected to contribute to future research on prompt generation. Another limitation pertained to the complexity of free-text notes, as the diagnosis was sometimes written in two different notes in distinct formats, with additional explanations that included target terms but not the periodontitis diagnosis. This complexity highlights the need for improved pre-processing techniques to extract meaningful information from such notes more effectively. Several minor limitations also existed, such as the potential to replace certain components with alternative packages or models to test performance. For future work, researchers could explore the use of ClinicalBERT, BioBERT, or other large language models as NER models to evaluate their effectiveness in this context. Furthermore, other GPT series models could be considered for prompt generation as an alternative to GPT-J.

In addition to addressing these limitations, future work could focus on expanding the scope of the study to include more diverse datasets and clinical conditions, ranging from rare diseases to common chronic conditions. This would enable researchers to assess the generalizability of the prompt generation approach across different domains and further demonstrate its potential in various healthcare settings. We are aware of the recent work of ChatAug,[21] which utilized GPT to generate synthetic notes to enhance the performance of local training models. Future research could build upon this concept by incorporating synthetic data generation techniques alongside prompt generation, exploring the synergistic effects of these methods on model performance. This could potentially lead to more robust, accurate, and efficient AI models that address various clinical challenges. Additionally, future work could investigate the effectiveness of integrating prompt generation with other NLP techniques, such as transfer learning and active learning, to further improve the performance of AI models in healthcare. By combining these methods, it may be possible to create more powerful AI-driven solutions that can adapt to new clinical scenarios rapidly, facilitating better patient care and decision-making processes.

**Conclusions**
In this study, the use of prompt generation for training NER models in the dental health field was explored as a promising solution to confirm periodontitis diagnoses recorded in point of care-documented clinical notes. The difficulty and time-consuming nature of identifying, extracting, and generating training data from the variety of terms and formats in clinical notes could be overcome with prompt generation. Combining different types of prompt generation can further improve the results, and this approach allows it to be extended to other diseases besides periodontitis. By increasing the complexity and incorporating rare medical terminologies, other large language models like GPT can be adopted in the pipeline to achieve satisfactory results.

**Acknowledgment**
XJ is CPRIT Scholar in Cancer Research (RR180012), and he was supported in part by Christopher Sarofim Family Professorship, UT Stars award, UTHealth startup, the National Institute of Health (NIH) under award number R01AG066749, R01LM013712, and U01TR002062, and the National Science Foundation (NSF) #2124789. This project is also approved by UTHealth IRB #HSC-DB-21-0616.